# CONCEPTUAL ASSOCIATION FOR COMPOUND NOUN ANALYSIS

**Abstract**

This paper describes research toward the automatic interpretation of compound nouns using corpus statistics. An initial study aimed at syntactic disambiguation is presented. Corpus derived lexical associations have proven successful for prepositional phrase attachment (Hindle and Rooth, 1993) suggesting that a similar approach may prove useful for compound noun analysis. The approach presented bases associations upon thesaurus categories rather than individual words, a technique described elsewhere as conceptual association (Resnik and Hearst, 1993). Association data is gathered from unambiguous cases extracted from a corpus and is then applied to the analysis of ambiguous compound nouns. While the work presented is still in progress, a first attempt to syntactically analyse a test set of 244 examples shows 75% correctness. Future work is aimed at improving this accuracy and extending the technique to assign semantic role information, thus producing a complete interpretation.

## INTRODUCTION

**Compound Nouns:** Compound nouns (CNs) are a commonly occurring construction in language consisting of a sequence of nouns, acting as a noun phrase; *fruit tree farmer,* for example. For a detailed linguistic theory of compound noun syntax and semantics, see Levi (1978). Compound nouns (also known as noun-noun compounds, complex nominals or noun sequences) are analysed syntactically by means of a rule such as N → N N which is applied recursively. Compounds of more than two nouns are ambiguous in syntactic structure. The first step in producing an interpretation of a CN is an analysis of the attachments within the compound. Syntactic parsers cannot choose an appropriate analysis, because attachments are not syntactically governed. Without semantic knowledge, multiple ambiguities arise, resulting in inefficient parsing. The current work presents a system for automatically deriving a syntactic analysis of arbitrary CNs in English using corpus statistics to provide lexical-semantic information.

**Task description:** The initial task can be formulated as choosing the most probable binary bracketing for a given noun sequence, known to form a compound noun, without knowledge of the context.

**Corpus Statistics:** The need for wide ranging lexical-semantic knowledge to support NLP, commonly referred to as the ACQUISITION PROBLEM, has generated a great deal of research investigating automatic means of acquiring such knowledge. Much work has employed carefully constructed parsing systems to extract knowledge from machine readable dictionaries (e.g., Vanderwende, 1993). These systems require significant knowledge to support their acquisition attempts. Other approaches have used rather simpler, statistical analyses of large corpora (for example, Church et al, 1991). Such work has established that corpus statistics are a promising source of automatic lexical-semantic knowledge.

Hindle and Rooth (1993) used a rough parser to extract lexical preferences for prepositional phrase (PP) attachment. The system counted occurrences of unambiguously attached PPs and used these to define LEXICAL ASSOCIATION between prepositions and the nouns and verbs they modified. This association data was then used to choose an appropriate attachment for ambiguous cases. When trained on a 13 million word corpus of news, the system could make correct attachments on nearly 80% of 880 test sentences. This was close to the performance of human subjects. Counting unambiguous cases in order to make inferences about ambiguous ones is a paradigm adopted in the current work. An explicit assumption is made that lexical preferences are relatively independent of the presence of syntactic ambiguity.

Subsequently, Hindle and Rooth's work has been extended by Resnik and Hearst (1993). Resnik and Hearst attempted to include information about typical prepositional objects in their association data. They introduced the notion of CONCEPTUAL ASSOCIATION in which associations are measured between groups of words considered to represent concepts, in contrast to single words. Such class-based approaches are used because they allow each observation to be generalized thus reducing the amount of data required. In the current work, a freely available version of Roget's thesaurus is used to provide the grouping of words into concepts, which then form the basis of conceptual association. The research can thus be seen as investigating the application of several key ideas in Hindle and Rooth (1993) and in Resnik and Hearst (1993) to the solution of an analogous problem, that of compound noun analysis. However, both these works were aimed solely at syntactic disambiguation. The goal of semantic interpretation remains to be investigated.

## METHOD

**Extraction Process:** The corpus used to collect information about compound nouns consists of some

7.8 million words from Grolier's multimedia on-line encyclopedia. The University of Pennsylvania morphological analyser provides a database of more than 315,000 inflected forms and their parts of speech. The Grolier's text was searched for consecutive words listed in the database as unambiguous nouns and separated only by white space. This prevented comma-separated lists and other non-compound noun sequences from being included. It also ensured that all words extracted were being used as nouns. However, it did eliminate many CNs from consideration because many nouns are occasionally used as verbs and are thus ambiguous for part of speech. This resulted in 35,974 noun sequences of which all but 655 were pairs. The first 1000 of the sequences were examined manually to check that they formed CNs. Only 2% were not compound nouns, thus establishing a reasonable utility for the extraction method. The pairs were then used as a training set, on the assumption that a two word noun compound is unambiguously bracketed [1].

**Thesaurus Categories:** The 1911 version of Roget's Thesaurus contains 1043 categories, with an average of 34 single word nouns in each. These categories were used to define concepts in the sense of Resnik and Hearst (1993). Each noun in the training set was tagged with a list of the categories in which it appeared [2]. All sequences containing nouns not listed in Roget's were discarded from the training set.

**Gathering Associations:** The remaining 24,285 pairs of category lists were then processed to find a conceptual association (CA) between every ordered pair of thesaurus categories (X, Y) using the following formula for CA (X, Y) where i and j range over all possible thesaurus categories:

Let **AMBIG**(w) = the number of thesaurus categories w appears in (the ambiguity of w).

Let **COUNT**($w_1$, $w_2$) = the number of instances of $w_1$ modifying $w_2$ in the training set

Let **FREQ**($t_1$, $t_2$) =
$$\sum_{w_1 \text{ in } t_1} \sum_{w_2 \text{ in } t_2} \frac{COUNT(w_1, w_2)}{AMBIG(w_1) \cdot AMBIG(w_2)}$$

Let **CA** ($t_1$, $t_2$) =
$$\frac{FREQ(t_1, t_2)}{\sum_{\forall i} FREQ(t_1, i) \cdot \sum_{\forall i} FREQ(i, t_2)}$$

Note that this measure is asymmetric. CA (X, Y) measures the tendency for X to modify Y in a compound noun, which is distinct from CA (Y, X).

**Automatic Compound Noun Analysis:** Given these associations, the following procedure can be used to syntactically analyse ambiguous CNs. Suppose the compound consists of three nouns: $w_1$ $w_2$ $w_3$. A left-branching analysis ($[[w_1\ w_2]\ w_3]$) indicates that $w_1$ modifies $w_2$, while a right-branching analysis ($[w_1\ [w_2\ w_3]]$) indicates that $w_1$ modifies something denoted primarily by $w_3$. For each $w_i$ (i = 2 or 3), choose categories $S_i$ (with $w_1$ in $S_i$) and $T_i$ (with $w_i$ in $T_i$) so that CA ($S_i$, $T_i$) is greatest. These categories represent the most significant possible word meanings for each possible attachment. Then choose $w_i$ so that CA ($S_i$, $T_i$) is maximum and bracket $w_1$ as a sibling of $w_i$. We have thus chosen the attachment having the most significant association in terms of mutual information between thesaurus categories.

In compounds longer than three nouns, this procedure can be generalised by selecting, from all possible bracketings, that for which the product of greatest conceptual associations is maximized.

## RESULTS

**Test Set and Evaluation:** Of the noun sequences extracted from Grolier's, 655 were more than two nouns in length and were thus ambiguous. Of these, 308 consisted only of nouns in Roget's and these formed the test set. All of them were triples. The remainder were discarded. Using the full context of each sequence in the test set, the author analysed each of these, assigning one of four possible outcomes. Some sequences were not CNs (as observed above for the extraction process) and were labeled E. Other sequences exhibited what Hindle and Rooth (1993) call SEMANTIC INDETERMINACY, where the meanings associated with two attachments cannot be distinguished in the context. These were labeled I. The remainder were labeled L or R depending on whether the actual analysis is left- or right-branching.

TABLE 1 - Test set analysis distribution:

| Labels | L | R | I | E | Total |
|---|---|---|---|---|---|
| Count | 163 | 81 | 35 | 29 | 308 |
| Fraction | 53% | 26% | 11% | 9% | 100% |

Proportion of different labels in the test set.

Table 1 shows the distribution of labels in the test set. Hereafter only those triples that received a bracketing (L or R) will be considered.

TABLE 2 - Results of test:

| x | Output Left | Output Right |
|---|---|---|
| Actual Left | 131 | 32 |
| Actual Right | 30 | 51 |

The proportions of correct and incorrect analyses.

---

[1] This introduces some additional noise, since extraction can not guarantee to produce complete noun compounds
[2] Some simple morphological rules were used at this point to reduce plural nouns to singular forms

The attachment procedure was then used to automatically assign an analysis to each sequence in the test set. The resulting correctness is shown in Table 2. The overall correctness is 75% on 244 examples. The results show more success with left branching attachments, so it may be possible to get better overall accuracy by introducing a bias.

## DISCUSSION

**Related Work:** While broad coverage compound noun analysis using lexical semantics has received little attention, there are two notable systems that are related to the current work. The most sophisticated is the SENS system described in Vanderwende (1993). SENS utilizes semantic features that are extracted from machine readable dictionaries by means of structural patterns applied to definitions. These features are then matched by heuristics which assign likelihood estimates to each possible semantic role. The result is a list of ranked interpretations (role assignments). The work only addresses the assignment of semantics to pairs of nouns and does not mention the problem of resolving syntactically ambiguous compounds.

The technique reported in Pustejovsky et al (1993) is aimed at bracketing ambiguous CNs and is far simpler. While attempting to extract taxonomic relationships, their system heuristically brackets CNs by searching elsewhere in the corpus for subcomponents of the compound. Such matching fails to take account of the natural frequency of the words and is likely to require a much larger corpus for accurate results. Unfortunately, they provide no evaluation of the performance afforded by their approach.

**Future Plans:** A useful side effect of employing thesaurus categories is that some sense disambiguation is performed. In *minority business development* the word *minority* is ambiguous for sense. When the analyser selects an attachment it also selects a category containing *minority* on which to base the decision. In this case, it correctly selects *Fewness (#103)* over *Inferiority (#34)* and *Youth (#127)*. In future work, an evaluation of the accuracy of these sense choices will be performed. A more sophisticated noun sequence extraction method should improve the results, providing more and cleaner training data.

Many sequences had to be discarded because they contained nouns not in the 1911 Roget's. A more comprehensive and consistent thesaurus needs to be used. An investigation of different association schemes is also planned, perhaps allowing for evidence from several categories to be combined.

Compound noun analyses often depend on contextual factors. Any analysis based solely on the static semantics of the nouns in the compound cannot account for these effects. The system presented in this paper is limited by this. However, it is not possible to tell from the work so far what a good performance level is. To establish a performance target an experiment is planned using human subjects, who will be given ambiguous noun compounds, out of context, and asked to choose attachments for them.

Finally, syntactic bracketing is only the first step in interpreting compound nouns. Once an attachment is established, a semantic role needs to be selected as is done in SENS. Given the promising results achieved for syntactic preferences, it seems likely that semantic preferences can also be extracted from corpora. This is the main area of ongoing research within the project.

## CONCLUSION

The current work uses thesaurus category associations gathered from an on-line encyclopedia to make analyses of compound nouns. An initial study of the syntactic disambiguation of 244 compound nouns has shown promising results, with an accuracy of 75%. Several enhancements are planned along with an experiment on human subjects to establish a performance target for systems based on static semantic analyses. The extension to semantic interpretation of compounds is the next step and represents promising unexplored territory for corpus statistics.